\documentclass{bmvc2k}

\usepackage{multirow}
\usepackage{pgfplots}
\usepackage{color,soul}

\title{Part Localization using Multi-Proposal Consensus for Fine-Grained Categorization}

\addauthor{Kevin J. Shih}{kjshih2@illinois.edu}{1}
\addauthor{Arun Mallya}{amallya2@illinois.edu}{1}
\addauthor{Saurabh Singh}{ss1@illinois.edu}{1}
\addauthor{Derek Hoiem}{dhoiem@illinois.edu}{1}

\addinstitution{
 University of Illinois\\
 Urbana-Champaign\\
 IL, US
}

\runninghead{Shih, Mallya, Singh, Hoiem}{Multi-Proposal Part Localization}

\def\etal{\emph{et al}\bmvaOneDot}

\def\argmin{\mathop{\rm argmin}}

\newcommand{\boldhead}[1]{\vspace{0.02in}\noindent\textbf{#1}}

\usepackage{wrapfig}

\begin{document}

  \maketitle


\begin{abstract}
We present a simple deep learning framework to simultaneously predict keypoint
locations and their respective visibilities and use those to achieve 
state-of-the-art performance for fine-grained classification. We show that by 
conditioning the predictions on object proposals with sufficient image support, 
our method can do well without complicated spatial reasoning. Instead, 
inference methods with robustness to outliers, yield state-of-the-art for keypoint localization.
We demonstrate the effectiveness of our accurate keypoint
localization and visibility prediction on the fine-grained bird recognition
task with and without ground truth bird bounding boxes, and 
outperform existing state-of-the-art methods by over 2\%. 
\end{abstract}


\section{Introduction}
\label{sec:intro}
Fine-grained image categorization is the task of accurately separating 
categories where the distinguishing features may be as minute as a different 
fur pattern, shorter horns, or a smaller beak. The widely accepted and popular 
approach of dealing with such a task is intuitive: align analogous 
regions and hone in on where you expect the differences to be. The analogous 
regions are usually defined by keypoints. Therefore, to perform well
one would require not only accurate object-level localization, but also part 
and/or keypoint localization.

For keypoint localization, the most common approach is to learn a set of 
keypoints detectors to model appearance and an associated spatial 
model~\cite{zhang2014part, liu2014part, liu2013bird, branson2014bird} to capture their spatial relations. 
Keypoint detectors generate a set of candidates and a spatial 
model is used to infer the most likely configuration.
Keypoint detectors typically model local appearance and 
thus an approach has to rely on expressive spatial models to capture long range 
dependencies. Alternatively, the keypoint detectors could condition their predictions
on larger spatial support and jointly predict several 
keypoints~\cite{bourdev2009}, then the need 
for expressive spatial models could be eliminated leading to simpler models.

For effective fine-grained category detection, the keypoint localization method must have high accuracy, low false 
positive rates, and low false negative rates. Missed or poorly localized predictions
make it impossible to extract the relevant features for the task at hand. If a keypoint is falsely determined to be present within a region, it is 
hard to guarantee that it will appear at a reasonable location. In the 
case of localizing keypoint-defined 
regions of an image, such as head or torso of a bird, a single outlier in the 
keypoint predictions can significantly distort the predicted area. This specific case is noteworthy as several of the current 
best-performing methods on the CUB 200-2011 birds dataset \cite{WahCUB_200_2011} rely on deep-network based features 
extracted from localized part regions 
\cite{zhang2014part,branson2014bird,liu2013bird,berg2013poof}.

In this work, we tackle the problem of learning a keypoint localization model 
that relies on larger spatial support to jointly localize several keypoints and 
predict their respective visibilities.
Leveraging recent developments in Convolutional Neural Networks (CNNs), we 
introduce a framework that outperforms the 
state-of-the-art on the CUB dataset. Further, while CNN-based methods suffer from a loss of image resolution due to the 
fixed-sized inputs of the networks, we introduce a simple sampling scheme that 
allows us to work around the issue without the need to train cascades of 
coarse-to-fine localization networks \cite{toshev2014deeppose,sun2013deep}.
Finally, we test our predicted keypoints on the fine-grained recognition task. 
Our keypoint predictions are able to significantly boost the performance of 
current top-performing methods on the CUB dataset. Our major contributions 
include: \vspace{-1mm}
\begin{enumerate}
	\item State-of-the-art keypoint and region (head, torso, body) 
localization with visibility prediction using a single neural network 
based on the AlexNet \cite{krizhevsky2012imagenet} architecture. \vspace{-2mm}

	\item A sampling scheme to significantly improve keypoint 
prediction performance without the use of cascades of coarse-to-fine 
localization networks \cite{toshev2014deeppose,sun2013deep}. \vspace{-2mm}

	\item Improvement of the state-of-the-art performance of 
\cite{zhang2014part} on the CUB classification task by using our predicted 
keypoints with significant gains when the 
groundtruth bird bounding box is not provided during test time.

\end{enumerate}


\section{Related Work}

\boldhead{Fine Grained Recognition:} Prior work focuses on localizing
informative parts of objects and then extracting features from them for
classification. Using pairs of localized keypoints, Berg \etal~\cite{berg2013poof} learn a set of highly discriminative features for
fine-grained classification. Farrell \etal~\cite{farrell2011birdlets}
and Branson \etal~\cite{branson2014bird} use pose normalized
representations of birds and their regions (head, torso, entire bird) followed
by feature extraction for classification. Liu \etal~\cite{liu2013bird}
extend the exemplar based model of \cite{belhumeur2011localizing} with pose
information for keypoint localization and subsequent classification of birds.
Based on the very successful framework of the RCNN \cite{girshick2014cvpr},
Zhang \etal~\cite{zhang2014part} perform bird classification using
three localized bird regions: head, torso, and full body.

The above mentioned methods are highly dependent on accurate keypoint and bird
region localization. In fact, \cite{berg2013poof,farrell2011birdlets} rely on
the groundtruth bird bounding box at test time to localize keypoints and to
perform classification. Our method overcomes this bottleneck of localization
and we demonstrate state-of-the-art classification performance using the framework of
\cite{zhang2014part} along with our localized regions. 

         
\boldhead{Object Region Proposals:}
Region proposals combined with deep network systems are an efficient solution for finding objects in an image. Recent works
use region proposals as initial object candidates to either reduce their
search space \cite{shihlearning, girshick2014rich} or to refine their
localization \cite{fidler2013bottom}. Instead of exhaustively sliding a
window feature extractor on an image at all locations, scales, and aspect
ratios, region proposal methods are used to quickly identify a smaller and
manageable set of image regions which have high recall for objects present in
the image. The time saved enables the use of more expensive feature extraction
and processing. Popular region proposal methods include
\cite{carreira2010constrained, EndresPAMI2014,cheng2014bing,
uijlings2013Selective, zitnick2014edge}. In our work, we use Edge
Boxes~\cite{zitnick2014edge} for its fast computational speed and its effective scoring method that allows us to
further reduce the number of candidates needed test time.

\boldhead{Pose Detection \& Regression with Deep Networks:} Our method for
keypoint localization mainly draws inspiration from the use of
regression in networks in the MultiBox approach by Erhan \etal~\cite{erhan2014scalable}. The authors train a
deep network which regresses a small number of bounding boxes ($\sim 100$) as
object bounding box proposals, along with a confidence value for each bounding
box.

Regression for localization of keypoints has previously been explored by Toshev
\etal~\cite{toshev2014deeppose}. They use a cascade of deep network based
regressors for human pose estimation to refine the keypoint predictions. At
each stage, the network uses a region around the previous prediction to acquire
higher resolution inputs and solve the fixed-resolution network input issue. In
contrast, our work relies on multiple regions sampled with Edge Boxes
from the image and simultaneously predicts all keypoints. Varying
sized regions provide varying resolution and context, and we achieve
more robust predictions from multiple regions with statistical outlier removal.

One of the closest works to ours on the CUB dataset is that of Liu
\etal~\cite{liu2013bird, liu2014part}. They achieve remarkable performance on both keypoint
localization and visibility prediction using ensembles of pose exemplars and
part-pair detectors.  We compare our performance with theirs
using metrics defined in their work.

In contemporary works, the Deep LAC model~\cite{lin2015deep} bridges a localization regression network and classification network to train simultaneously to perform on similar tasks to our own. While their setup is very similar to our own, they directly target the localization of entire head and torso boxes whereas we target the keypoints that define said boxes. We include their accuracies for comparison in the localization and recognition experiments.


\section{Method}
\label{sec:method}
We design our model to simultaneously predict keypoint locations and 
their visibilities for a given image patch. To share the information across 
categories, our model is trained in a category agnostic manner. At test time, 
we efficiently sample each image with Edge Boxes, make predictions from each Edge Box, 
and reach a consensus by thresholding for visibility and reporting the medoid.

\subsection{Training Convolutional Neural Networks for Keypoint Regression}

Our network is based on AlexNet~\cite{krizhevsky2012imagenet}, but modified to 
simultaneously predict all keypoint locations and their visibilities for any 
given image patch. AlexNet is an architecture with 5 convolutional layers 
and 3 fully connected layers. Henceforth, we refer to the 3 fully connected 
layers as fc6, fc7, and fc8. We replace the final fc8 layer with two separate 
output layers for keypoint localization and visibility respectively. Our network 
is trained on Edge Box \cite{zitnick2014edge} crops
extracted from each image and is initialized with a pre-trained AlexNet
\cite{krizhevsky2012imagenet} trained on the ImageNet \cite{imagenet} dataset.
Each Edge Box is warped to 227$\times$227 pixels before it can be fed through
the network. We apply padding to each Edge Box such that the warped
227$\times$227 pixel crop has 16 pixels of buffer in each direction.

Given $N$ keypoints of interest, we train a network to output an $N$
dimensional vector $\hat{v}$ and a $2N$ dimensional vector $\hat{l}$
corresponding to the visibility and location estimates of each of the keypoints $k_i$,
$i\in\{1, N\}$, respectively. The corresponding groundtruth targets
during training are $v$ and $l$. We define $v$ to
consist of indicator variables $v_i\in \{0,1\}$ such that $v_i = 1$ if
keypoint $k_i$ is visible in the given Edge Box image before padding is
performed, and 0 otherwise. The groundtruth location vector $l$ is of
length $2N$ and consists of pairs $(l_{x_i}, l_{y_i})$ which are the normalized
$(\tilde{x},\tilde{y})$ coordinates of keypoint $k_i$ with respect to the
un-padded Edge Box image. Output predicted from the network,
$\hat{v}_i\in[0,1]$, acts as a measure of confidence of keypoint
visibility, and 2D locations predicted by the network are denoted by
$\hat{l}_i$.


We use the \emph{Caffe} framework \cite{jia2014caffe} for training our deep
networks. To train a network optimized for both tasks simultaneously, we define
our losses as follows:
\begin{align}
  \mathcal{L}_{vis} = ||v - \hat{v}||^2_2 \quad\text{and}\quad
&\mathcal{L}_{loc} = \sum_{i=1}^{N} v_i \cdot \left [(l_{x_i} -
\hat{l}_{x_i})^2 + (l_{y_i} - \hat{l}_{y_i})^2\right ] \\ &\mathcal{L}_{net} =
\mathcal{L}_{vis} + \mathcal{L}_{loc}
\end{align}

The visibility loss $\mathcal{L}_{vis}$ is the squared Euclidean distance
between the ground truth visibility label vector $v$, and the predicted
visibility vector $\hat{v}$. The values in our $\hat{v}$'s always lie between 0
and 1 as they are obtained after squashing network outputs with a sigmoid
function. The keypoint localization loss $\mathcal{L}_{loc}$ is a modified
Euclidean loss, in which we set the loss between the prediction and the target
to be 0 if $v_i=0$ i.e. if the keypoint $k_i$ is absent in the given image. The
final training loss ($\mathcal{L}_{net}$) is given by the sum of the two
losses.

To construct our training set for predicting keypoint visibility and locations,
we extract up to 3000 Edge Boxes per image. To train a robust predictor, we
need a collection of training images with high variability in which different
subsets of keypoints are visible. We generate examples that satisfy this
criteria by retaining a subset of Edge Boxes which have at least 50\% of their
area contained inside the groundtruth bounding box and have at least 20\%
intersection over union overlap (IOU) with the groundtruth bounding box. We
also included up to 50 random boxes per image from outside the bounding box as
negative background examples. We augment our dataset with left/right flips. After
flipping, appropriate changes were applied to the label vectors. This consisted
of swapping orientation-sensitive keypoints such as ``left eye'' and ``left
wing'' with ``right eye'' and ``right wing'', and updating their respective
coordinates and visibility indicators. We first train our model on 25 images
per class and tune algorithmic and learning rate
parameters on a held-out validation set comprising the remaining 4-5
images per class. Finally, we re-train using the entire training set
before running our model on the test set.

\begin{figure*}
	\centering
	\includegraphics[width=0.8\textwidth]{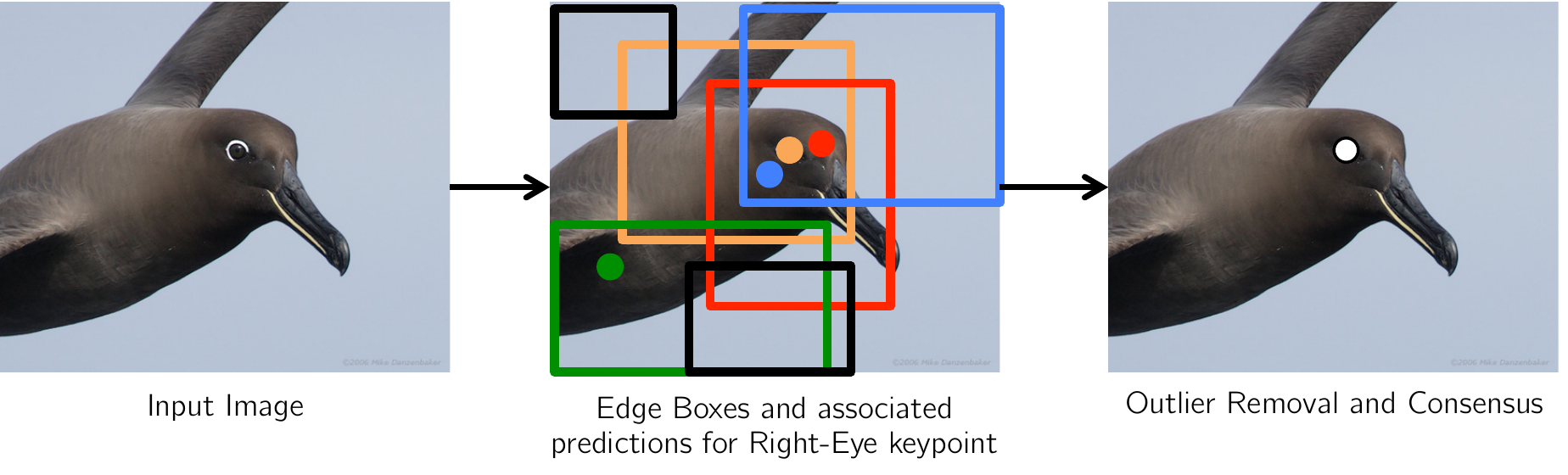}
  \caption{\small The pipeline of our keypoint localization process: Given an
input image, we extract multiple edge boxes. Using each edge box, we make
predictions for the location of each of the 15 keypoints, along with their
visibility confidences. We then find the best predicted location by performing
confidence thresholding and finding the medoid. The process is illustrated for
the right eye keypoint (Black edge boxes without associated dots make
predictions with confidences below the set threshold, and green is an outlier
with a high confidence score). } \label{fig:pipeline}
\end{figure*}

\subsection{Combining Multiple Keypoint Predictions}
Our algorithm for dealing with predictions from multiple Edge Boxes at test time
is illustrated in Fig.~\ref{fig:pipeline}. Due to the variance from making
predictions from multiple unique subcrops of the image, we need to form a consensus from the multiple predictions. In our experiments, we found
that after removing predictions with low visibility confidences, the remaining
predictions had a peaky distribution around the ground truth. We use medoid as
a robust estimator for this peak and found it to be effective in most cases 
(Fig.~\ref{fig:qual_res_kpts}).
For the task of localizing part regions around keypoints (described in section
\ref{ss:classification}), we found on our train/val split that we achieved
better localization performance if we kept a set of good predictions (referred
to as \emph{inliers}) instead of using only the medoid. We now describe our
procedure for obtaining a tight set of inliers and our choice of parameters.
For the keypoint prediction task, we only use the visibility thresholds and
report the medoid.\\

\vspace{-10pt}

\boldhead{Case 1: Ground Truth Object Box Given:} We first describe our method
in the case that the ground truth object boxes are given. Using the ground
truth object box, we retain the generated Edge Boxes that are mostly contained
within and have an IOU of at least 0.2. This results in roughly 50-200 remaining
Edge Box subcrops per image. Each subcrop is then independently fed through our
keypoint prediction network, returning a set of normalized keypoint predictions
and visibilities.

Because each subcrop is expected to cover less than the whole object
and contain only a subset of the keypoint predictions, we drop any prediction 
if its corresponding visibility is below 0.6. Because we make use of multiple
overlapping subcrops, it is very likely that at least one of them will lead to
a prediction with a sufficiently high visibility score, thereby allowing us to
be much more aggressive with the false positive filtering.

Given multiple remaining keypoint predictions per keypoint with sufficiently
high visibility scores, we then proceed to remove outliers. To do so, we
threshold on a modified Z-score based on a description given by Iglewicz and
Hoaglin \cite{iglewicz1993detect}. The modified $Z$-score is one that is
re-defined using medoids and medians in place of means, as the former estimates
are more robust to outliers.

Let $p_i$ where $i = 1,\cdots,M$ be the set of $M$ surviving  un-normalized keypoint predictions (for a given keypoint) in $(x,y)$ image coordinates. We first define
$\bar{p}$ to be the medoid prediction such that: \begin{equation}
  \bar{p} = \argmin_{p_j}\sum_{i=1}^{M}||p_j-p_i||_2,~ j \in \{1,...,M\}
\end{equation}

In other words, $\bar{p}$ is the prediction such that its Euclidean distance
from all other predictions for that keypoint is minimal. While this optimization is costly at a
large scale, we typically deal with only 10-20 predictions at a time after thresholding for visibility scores. To
compute the modified $Z$-score we use: \begin{equation}
  Z_i = \frac{\lambda\ ||p_i - \bar{p}||_2}{\text{median}\ (||p_i - \bar{p}||_2)},~ i \in \{1,...,M\}
\end{equation}

Here, the denominator is the median absolute deviation, or simply the median
distance from the medoid $\bar{p}$. We use the recommended $\lambda = 0.6745$.
The above procedure is separately computed for all 15 sets of keypoint
prediction candidates. Finally, we drop any keypoint prediction with $Z_i >
0.35$, a threshold that was experimentally determined on the held-out set.\\

\boldhead{Case 2: Ground Truth Object Box Not Given:} Our ground truth object box not
given scenario requires little change from the above case. Using the Edge Box
ranking, we found that most of our ``good'' Edge Boxes fell within the top 600
Edge Boxes per image, saving us a lot of computation. Tuning parameters on our
train/val split, we found that an even more aggressive visibility threshold of
0.94 and a Z-score threshold of 0.3 gave the best results.\\

\boldhead{Medoid-Shift:} While the simple Z-score thresholding combined with
the medoid achieves excellent results, as we will demonstrate in the results
section, we were able to further improve our results by using
medoid-shifts~\cite{sheikh2007mode}. We use the medoid of the largest output
cluster from the algorithm instead of the medoid computed over all the
visibility-filtered predictions. 

\subsection{Bird Classification} \label{ss:classification}
We verify the effectiveness of our localized parts by implementing the simple
classification framework as described in \cite{zhang2014part}. Using the
keypoints, three regions are identified from each bird: head, torso, and whole
body. The head is defined as the tightest box surround the beak, crown,
forehead, eyes, nape, and throat. Similarly, the torso is the box around the
back, breast, wings, tail, throat, belly, and legs. The whole body bounding box
is the object bounding box provided in the annotations. To perform
classification, fc6 features are extracted from these localized regions,
concatenated into a feature vector of length $4096\times3$, and used for
200-way linear 1-vs-all SVM classification.

To handle the case when ground truth bounding box is not given at test
time, we use an overlap heuristic based on the predicted head and torso boxes. We
first start by finding the tightest box around the predicted head and torso
boxes. While this initial box will do
well for birds in their canonical poses, it will
result in an undersized box in many cases because the keypoints do not always capture
the full extent of the bird. We then assume that there exists an Edge Box with a high edge score
that better captures the whole bird.  To let the box expand to capture more of
the object, we first identify the Edge Boxes such that the tightest box is at
least 90\% contained within and has at least 50\% IOU overlap. The final whole
body bounding box is the Edge Box that passes both criteria that also has the
highest Edge Box object score. If no Edge Box passes the overlap test, we
fall back to the starting tightest box.


\section{Experiments and Results}
We evaluate our prediction model on the challenging Caltech-UCSD Birds
dataset~\cite{WahCUB_200_2011}. This dataset contains 200 bird categories with
15 keypoint location and visibility labels for each of the total of 11788 images. We first evaluate our
keypoint localization and visibility predictions against other top-performing
methods. Next, we demonstrate their effectiveness in the fine-grained
categorization task by significantly improving state-of-the-art through
better localization. 

\subsection{Keypoint Localization and Visibility Prediction}

\begin{figure*}[h!]
	\centering
	\includegraphics[width=0.8\textwidth]{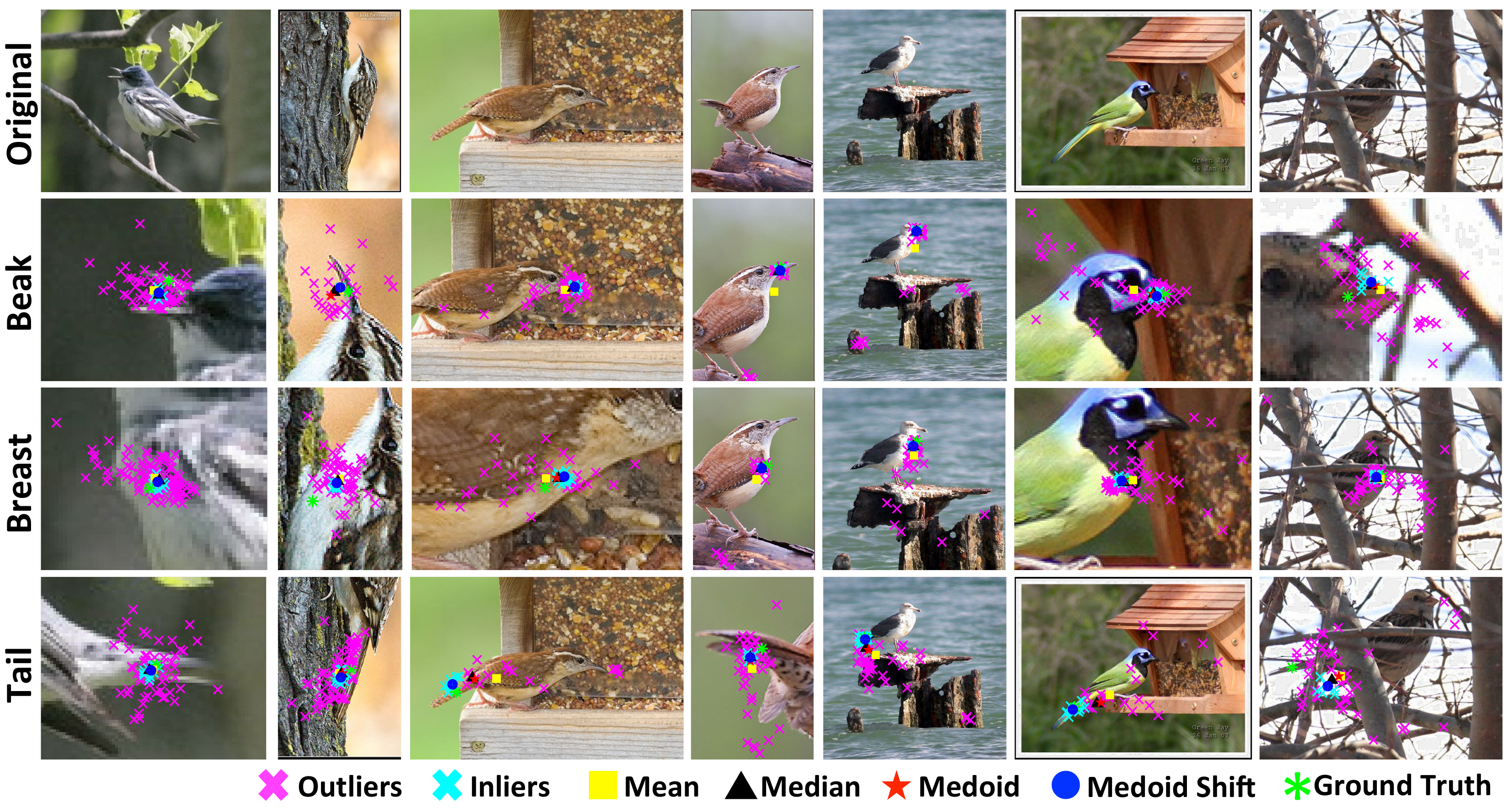}
  \caption{\small Qualitative results for a subset of the keypoints.
Predictions for most of the images cluster tightly. Therefore, simple
prediction methods such as medoids work well. Medoid shift adds to
   the robustness, leading to further improvements (second last column).
   Primary failure mode is when visibility thresholding fails to rule out clusters of false positives (bottom right).} 
\label{fig:qual_res_kpts}
\end{figure*}

Table~\ref{tbl:locandvis} reports our keypoint and visibility performance
without using any ground truth bounding box information.
Our medoid method reports the medoid of predictions above a visibility
threshold, as seen in the red star in Fig.~\ref{fig:qual_res_kpts}. Our ``mdshift'' 
method reports the new medoid computed using
medoid-shift, which is the blue circle in Fig.~\ref{fig:qual_res_kpts}. 
We used the evaluation code provided by the authors of
\cite{liu2013bird} to measure our performance using the metrics defined in
their work. In short, PCP (Percent Correct Parts) is the percentage of
keypoints localized within 1.5 times the annotator standard
deviation. We received the pre-computed standard deviatons and
evaluation code from
the authors of \cite{liu2013bird} to avoid any discrepancies during
evaluation. AE
(Average Error) is the mean euclidean prediction error, capped at 5 pixels,
computed across examples where a prediction was made and a ground truth
location exists. FVR and FIR refer to False Visibility Rate and False
Invisibility Rate respectively. The additional methods for comparison are the same as
listed in their paper.

\begin{wraptable}{R}{0.5\textwidth}  
\footnotesize
  \begin{center}
  \begin{tabular}{l|l|l|l|l}
    Method & PCP & AE & FVR & FIR\\
    \hline
    Poselets \cite{bourdev2010} & 24.47 & 2.89 & 47.9 & 17.15 \\
    Consensus \cite{belhumeur2011localizing} & 48.70 & 2.13 & 43.9 & 6.72\\
    Exemplar \cite{liu2013bird} & 59.74 & 1.80 & 28.48 & {\bf 4.52} \\
    \hline
    Ours (medoid)& 68.7 & 1.4 & {\bf  17.1} & 5.2\\
    Ours (mdshift) & {\bf 69.1} & {\bf 1.39} & {\bf 17.1} & 5.2\\
    \hline
    Human \cite{liu2013bird} & 84.72 & 1.00 & 20.72 & 6.03\\
  \end{tabular}
  \end{center}
  \caption{\small Localization and Visibility Prediction Performance of various methods without using the ground truth Bounding Box}
  \label{tbl:locandvis}
\end{wraptable}

Compared to the top-performing methods that also predict visibility, our method
achieves the best numbers in three out of four metrics. 
Our PCP and AE metrics outperform other methods in the table, with our
medoid-shift variant performing slightly better. Our FIR is
higher because we are using
the visibility threshold tuned on the part-localization task. A
slightly lowered threshold would lower the FIR and raise the FVR without
significantly affecting the PCP.

The highest reported PCP is 66.7\% due to Liu \etal~\cite{liu2014part}, which also predicts
visibilities but did not report them. We compare against
their PCP in Table~\ref{tbl:prtpcp}. Because our method differs
significantly from theirs, we outperform them
in only 7 of the listed part categories despite having a better overall PCP,
suggesting further improvements by targeting the differences in our models' behaviors.

\begin{table}
    \footnotesize
  \centering
  \begin{tabular}{l|l|l|l|l|l|l|l|l|l|l|l|l|l}

    PCP & Ba & Bk & Be & Br & Cr
        & Fh & Ey & Le & Wi & Na
        & Ta & Th & Total\\
    \hline
    \cite{liu2013bird} & 62.1& 49.0 & 69.0 & 67.0 & 72.9
        & 58.5 & 55.7 & 40.7 & 71.6 & 70.8
        & 40.2 & 70.8 & 59.7\\
    \cite{liu2014part} & 64.5 & {\bf 61.2} & 71.7 & 70.5 & 76.8
        & {\bf 72.0} & {\bf 70.0} & 45.0 & 74.4 & {\bf 79.3} 
        & 46.2 & {\bf 80.0} & 66.7 \\
    \hline
    Ours & {\bf 74.9}  & 51.8 & {\bf 81.8} & {\bf 77.8} & {\bf 77.7}
        & 67.5 & 61.3 & {\bf 52.9} & {\bf 81.3} & 76.1
        & {\bf 59.2} & 78.7 & {\bf 69.1} \vspace{3mm}

  \end{tabular}
  \caption{\small Comparison of per-part PCP with Liu \etal~\cite{liu2013bird,liu2014part}. The abbreviated part names from left
to right stand for back, beak, belly, breast, crown, forehead, eye, leg, wing, nape,
tail, and throat.}
\label{tbl:prtpcp}
\end{table}
  
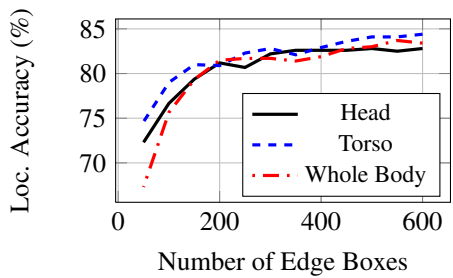
\begin{wrapfigure}{R}{0.5\textwidth}
  \begin{tikzpicture}
    \pgfplotstableread{data/localization-boxes-data.dat}
    \datatable
    \begin{axis}[
        height=4cm,
        width=6cm,
        xlabel=Number of Edge Boxes,
        ylabel=Loc. Accuracy (\%),
        grid=major,
        legend columns = 1,
        legend style={font=\small},
        legend entries={Head,Torso,Whole Body},        
        legend pos = south east]
      \addplot+[color=black,mark=none,mark size=2pt, very thick] table[x=boxes, y=H]
      from \datatable;
      \addplot+[color=blue,mark=none, mark size = 2pt,dashed, very thick] table[x=boxes, y=T]
      from \datatable;
      \addplot+[color=red,mark=none, mark size = 2pt, dash pattern=on 1pt off 3pt on 6pt off 3pt, very thick] table[x=boxes, y=B]
      from \datatable;
      \end{axis}
  \end{tikzpicture}
  \caption{\small Localization performance on our validation set while varying
the number of top Edge Boxes used. We only rely on the Edge Boxes as a means of
efficient sampling of the image, so our performance is barely affected by the
loss of well-localized boxes. }
  \label{fig:locbox}
\end{wrapfigure}

\subsection{Head and Torso Localization}
\begin{figure}
  \centering
  \includegraphics[width=0.9\paperwidth]{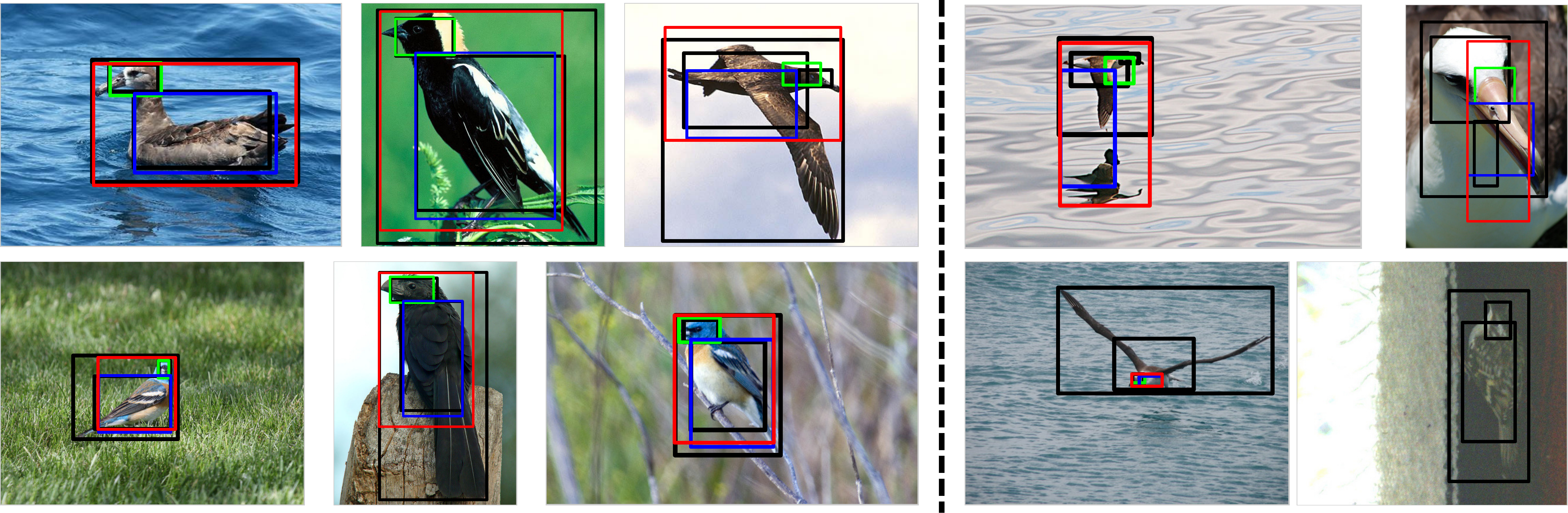}
  \caption{\small Examples of good (left) and failed (right) localization results:
The ground truth boxes are in solid black. The head, torso, and whole body
boxes are in green, blue and red respectively. The head is correctly localized
in most of the above examples. In the top row middle example, even
though the whole body box IOU is low, most of the missed area is actually
background due to the bird extending its wings. In the bad examples, we show
that we mostly fail in rare close-ups and when there are multiple instances.}
\label{fig:examples}
\end{figure}

We evaluate our ability to localize keypoint-defined part-regions on the test
set. Example predictions can be seen in Fig.~\ref{fig:examples}. In Table~\ref{Tbl:partloc}, we compare our part-localization accuracies
with other methods. We also compare with the simple case where we make
predictions by feeding just the ground truth boxes through the network
(single GT Bbox). (We also tried re-training a network on just GT
bounding boxes for this case, but it didn't perform as well.) Unlike
the keypoint prediction task, we retain a set of inliers after $Z$-score
thresholding to determine the extent of each part box. This was determined to
perform best on our validation set. The reported metrics are the percentage of
heads, torsos, and whole body boxes that were correctly localized with a >50\%
IOU. The ability to perform competitively on this task should correlate with a
high PCP and low FVR and FIR.

The results in Table~\ref{Tbl:partloc} demonstrate that our keypoint predictions are
useful in generating accurate part boxes. Our lower performing single GT Bbox method
suggests that our use of multiple predictions from Edge Boxes allows for more
accurate predictions. Further, we also computed head and torso boxes using the
keypoint predictions from Liu \etal~\cite{liu2013bird} as shown in the
``Exemplar'' row. Based on their accuracy, their
boxes should also be able to improve the results of ~\cite{zhang2014part}.

In Fig.~\ref{fig:locbox}, we also look at how our localization ability is
affected by the number of top Edge Boxes sampled from the image. As we
previously noted, the Edge Box edge scoring is effective enough that most
of the sufficiently well localized boxes we used in the ground-truth bounding
box given case fell within the top 600. However, as our model predicts
individual keypoints and visibilities, it does not need a well localized box at
test time at all. It merely needs a set of Edge Boxes that, combined, provide
enough coverage over the actual bird for it to predict keypoint locations and
visibilities. As such, our model is able to continue to localize over 70\% of
the head and torso boxes with at least 50\% IOU as the number of Edge Boxes
drops to 50. While the 50\% IOU recall of Edge Boxes for head and torsos on the
validation set were 66.36\% and 95.12\% at 600 boxes and 17.28\% and 62.20\%
respectively at 50 boxes, we demonstrate that we were able to localize these
parts with higher accuracy than would have been achievable had we used an
RCNN-based approach and tried to map Edge Boxes to the parts. 

\begin{table}[t!]
  \footnotesize
\begin{center}
\begin{tabular}{l|l|l|l|l}
   & Method & Head & Torso & Whole Body\\
  \hline
  \multirow{5}{*}{GT Bbox}
  & Part-Based RCNN \cite{zhang2014part} & 68.2 & 79.8 & N/A\\
  & Deep LAC \cite{lin2015deep} &74.0 & \bf{96.0} & N/A\\ \cline{2-5}
  & Ours (single GT bbox)& 75.6 & 90.2& N/A\\
  & Ours (multiple) & 88.8 & 93.9& N/A\\
  & Ours (multiple, mdshift) & \bf{88.9}&94.3& N/A\\
  \hline
  \multirow{5}{*}{No GT Bbox}
  & Part-Based RCNN \cite{zhang2014part} & 61.4 & 70.7& {\bf 88.3}\\
  & Exemplar \cite{liu2013bird} & 79.9 & 78.3 & N/A\\ \cline{2-5}
  & Ours (multiple) & 87.8 & {\bf 89.0} & 84.5\\
  & Ours (multiple, mdshift) & {\bf 88.0}&88.7 & 84.6\\
  \end{tabular}
\end{center}
\caption{\small Comparison of Part Localization Performance: Our method based
on keypoint prediction from Edge Boxes shows significant improvement over
previous work.} \label{Tbl:partloc}
\end{table}

\vspace{-5mm}

\subsection{Fine-Grained Classification}
We now test our part-predictions in a fine-grained classification
setting. These results are shown in the right half of
Fig. \ref{Fig:classification}. To do this, we train three networks to re-implement the three-part framework as
Zhang \etal \cite{zhang2014part} as described in section \ref{ss:classification}. The
oracle performance refers to the classification assuming ground truth keypoints
at test time. While Zhang \etal \cite{zhang2014part} reports an oracle accuracy of
82.0\%, we compare with the highest we were able to achieve with our implementation: 81.5\%. This is likely due to minor
differences in network training parameters. We also tried both fc6 and fc7 features and
found that fc6 performed a little better. Although Zhang \etal \cite{zhang2014part} and
Branson \etal \cite{branson2014bird} noted that their drops in accuracy from using ground
truth parts to predicted parts were surprisingly small, our
relative improvements suggest that it is still worthwhile to focus on
better localization. Further, we perform at least as well as the
contemporary Deep LAC model~\cite{lin2015deep}, likely due
to our better localization of the more discriminative head regions.

In the left half of Fig. \ref{Fig:classification}, we show how our accuracy is
affected from the ground truth keypoint ideal case (Oracle) to the use of
predicted keypoints (GT Bbox), and finally with the GT Bbox removed (No GT
Bbox). Unsurprisingly, the better localization at test time allows for a
significantly smaller drop as annotations are removed.

The same plot also shows an ablation test of individual parts. It appears that
the bulk of our performance comes from discriminating localized bird heads.
This is also supported by \cite{branson2014bird} which observed that of their
learned poses, the one that corresponded to the head was the most
discriminative. This suggests that most of our improvement over our base method
of \cite{zhang2014part} comes from significantly improving our head part
localization (shown in Table \ref{Tbl:partloc}).

\vspace{-3mm}

\pgfplotsset{
    /pgfplots/ybar legend/.style={
    /pgfplots/legend image code/.code={%
       \draw[##1,/tikz/.cd,yshift=-0.25em]
        (0cm,0cm) rectangle (3pt,0.8em);},
   },
}

\begin{figure}[ht]
  \centering
  \hspace{-5mm}
  \begin{minipage}{.55\textwidth}
 \begin{tikzpicture}
    \small
    \begin{axis}[        
        height=4cm,
        width = \textwidth,
        ybar,
        enlargelimits=0.15,
        legend style={font=\footnotesize, column sep=0.2cm, at={(0.5,-0.25)},anchor=north, legend columns=-1},
        legend cell align=left,
        ylabel={Class Acc. (\%)},
        xlabel={Parts Used},
        symbolic x coords={H+T+B,H+T,H,T},
        xtick=data,
        nodes near coords,
        every node near coord/.append style={font=\tiny},
        nodes near coords align={vertical},
        ylabel near ticks,
        bar width = 0.33cm
      ]
      \addplot coordinates {(H+T+B,81.5) (H+T,81.0) (H,71.5)  (T,61.0)};
      \addplot coordinates {(H+T+B,80.3) (H+T,78.8) (H,68.3)  (T,60.4)};
      \addplot coordinates {(H+T+B,78.3) (H+T,77.7) (H,68.2)  (T,58.7)};
      \legend{Oracle, GT Bbox, No GT Bbox}
    \end{axis}
 \end{tikzpicture}
  \label{fig:accuracy_comparison}
  \end{minipage}
  \hspace{-4mm}
  \begin{minipage}{.43\textwidth}
    \footnotesize
    \centering
    \begin{tabular}{l|l|l}
        & Method & Acc.\\
        \hline
        Oracle  & Oracle Parts + SVM & {\bf 81.5} \\ \hline
        \multirow{8}{*}{GT Bbox} & DPD \cite{zhang2013deformable} & 51.0\\
        & Symbiotic \cite{chai2013symbiotic} & 59.4\\
        & Alignment \cite{gavves2013fine} & 62.7\\
        & DeCAF \cite{donahue2013decaf} & 65.0\\
        & POOF \cite{berg2013poof} & 56.8\\
        & Part-Based RCNN \cite{zhang2014part} & 76.4\\
        & Deep LAC \cite{lin2015deep} & 80.3 \\
        \cline{2-3}
        & Ours (mult, medoid) & 80.3 \\
        & Ours (mult, mdshft) & {\bf 80.3} \\
        \hline
        \multirow{4}{*}{No GT Bbox} & Pose Norm \cite{branson2014bird} & 75.7\\
        & Part-Based RCNN \cite{zhang2014part} & 73.9\\ \cline{2-3}
        & Ours (mult, medoid) & 78.2\\
        & Ours (mult, mdshft) & {\bf 78.3}\\
      \end{tabular}      
      \label{Tbl:accuracy}
  \end{minipage}
  \vspace{2mm}
  \caption{\small On the left we show a comparison of classification
accuracies obtained using combinations of parts localized under different conditions (H: Head, T: Torso, B: Whole Body).
On the right, we compare our classification accuracy with other works.} 
  \label{Fig:classification}
\end{figure}
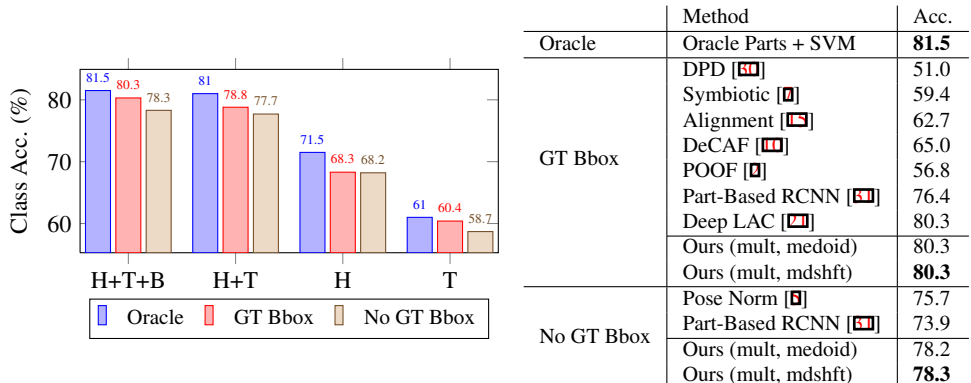

  \section{Conclusion} We presented a method for obtaining
state-of-the-art keypoint predictions on the CUB dataset. 
We demonstrated that conditioning the predictions on multiple object proposals
for sufficient image support can reliably improve localization predictions
without using a cascade of coarse-to-fine networks. We tackle the
problem of fixed-size inputs when using neural networks by sampling
predictions from several boxes and
determining the ``peak'' of the predictions with medoids. We then use our
predictions to significantly improve state-of-the-art methods on the
fine-grained classification task on the CUB dataset. In future work,
we intend to apply this method to human datasets as
well as to combine it with more sophisticated inference methods to deal
with multiple instances.

  \section{Acknowledgements}
This work is supported by NSF CAREER awards 1053768 and 1228082, NSF Award IIS-1029035, ONR MURI Award N00014-10-10934 and the Sloan Fellowship. In addition, we gratefully acknowledge the support of NVIDIA Corporation with the donation of the Tesla K40 GPUs used for this research.

  \bibliography{egbib}

\end{document}